\documentclass[letterpaper, 10 pt, conference]{ieeeconf}  

\IEEEoverridecommandlockouts                              

\usepackage{graphicx}
\usepackage{amsmath}
\usepackage{float}
\usepackage{commath}
\usepackage{subfig}
\usepackage{url}

\usepackage{ifthen}
\usepackage{bbm}
\usepackage{dsfont}
\usepackage{amssymb}
\usepackage{mathbbol}
\usepackage{multirow}
\usepackage{url}
\usepackage{breakurl}
\usepackage[breaklinks]{hyperref}
\usepackage{romannum}
\usepackage{caption,subcaption}
\usepackage{comment}
\usepackage{wrapfig}
\usepackage{txfonts}
\usepackage[english]{babel}

\title{\LARGE \bf
Reinforcement Learning Driven Cooperative Ball Balance in Rigidly Coupled Drones*
}

\author{Shraddha Barawkar$^{1}$ and Nikhil Chopra$^{2}$
\thanks{*This work was supported by the Maryland Robotics Center, University of Maryland}
\thanks{$^{1}$Shraddha Barawkar is with Maryland Robotics Center, University of Maryland, College Park, MD 20742, USA
        {\tt\small sbarawka@umd.edu}}%
\thanks{$^{2}$Nikhil Chopra is with the Department of Mechanical Engineering, University of Maryland,
        College Park, MD 20742, USA
        {\tt\small nchopra@umd.edu}}%
}

\begin{document}

\maketitle
\thispagestyle{empty}
\pagestyle{empty}

\begin{abstract}
Multi-drone cooperative transport (CT) problem has been widely studied in the literature. However, limited work exists on control of such systems in the presence of time-varying uncertainties, such as the time-varying center of gravity (CG). This paper presents a leader-follower approach for the control of a multi-drone CT system with time-varying CG. The leader uses a traditional Proportional-Integral-Derivative (PID) controller, and in contrast, the follower uses a deep reinforcement learning (RL) controller using only local information and minimal leader information. Extensive simulation results are presented, showing the effectiveness of the proposed method over a previously developed adaptive controller and for variations in the mass of the objects being transported and CG speeds. Preliminary experimental work also demonstrates ball balance (depicting moving CG) on a stick/rod lifted by two Crazyflie drones cooperatively. 
\end{abstract}

\section{INTRODUCTION}
Motivated by challenging control problems such as ball balance on a plate \cite{zarzycki2021fast}, in this paper, we study the ball balance capabilities using drone platforms.  Decentralized ball balance or handling of continuously moving CG in multi-drone CT systems using drones and RL is presented in this paper. Examples of real-world applications that motivate this work include a multi-drone flying car and heavy package delivery applications. The motion of people inside the flying vehicle and the movement of package contents (for package delivery) may continuously change the CG of the system. The next motivating factor for this research is the increasing use of machine learning-based closed-loop controllers in robotic systems. Reinforcement learning-based controllers may not require knowledge of system dynamics but are data-intensive and may not be endowed with closed-loop stability and performance guarantees. On the other hand, traditional control approaches can provide stability guarantees but rely on accurate system dynamics or at least a knowledge of the underlying physics. In this paper, we study the interaction of a machine learning-driven drone with a PID-controlled drone in the problem of decentralized ball balancing, which, to the best of our knowledge, has not been accomplished earlier in the literature.

The problem of time-varying CG in multi-drone CT systems can be addressed using adaptive control theory \cite{aastrom1983theory} based methods. However, such algorithms critically rely on the structure of the unknown dynamics and typically work well for constant unknown parameters. The performance limitations of this method can be seen~\cite{barawkar2024decentralized} from the oscillatory system waypoint navigation due to the delay in estimating time-varying CG in a multi-drone CT system. We next discuss the state-of-the-art of multi-robot/drone CT and deep RL.\\
Vision, admittance, and geometric control are a few of the diverse controllers used in the literature demonstrating multi-drone CT \cite{loianno2017cooperative, barawkar2023force, sreenath2013dynamics}. In handling system uncertainties in multi-drone/robot CT, \cite{culbertson2018decentralized,kawasaki2003adaptive} demonstrate decentralized adaptive control schemes for multi-industrial robot CT. Adaptive controller for cabled CT with multiple catenary robots is shown in \cite{cardona2021adaptive}. References \cite{aghdam2016cooperative, arab2021cooperative} present cabled multi-drone CT with offset and unknown fixed CG. The work of \cite{pierri2020cooperative} shows multi-drone CT of an unknown object with unknown CG using omnidirectional drones. However, the controller has not been tested for continuous time-varying CG in all these works. These works also do not consider rigid connections between the drones and the transported objects, wherein the rigid connections couple the orientation of drones and the object, thereby complicating the closed-loop control due to the coupled nonlinear dynamics. Additionally, addressing the impact of a time-varying CG is a complex challenge. 

In \cite{barawkar2019cooperative}, a centralized PID control was demonstrated on a multi-drone CT system in the presence of an unknown CG of the object. It should be noted that even if such a decentralized controller was developed, a decentralized multi-drone PID system may not be robust to GPS localization errors, especially in an outdoor environment. The reader is referred to \cite{barawkar2023force} for additional details.

Deep RL has been shown to perform better than humans in competitive games such as Chess, Atari, and Go. These advances using deep RL have recently been reported in champion-level drone racing \cite{kaufmann2023champion}. Deep RL has also been implemented for robotic manipulation\cite{haarnoja2018composable}, quadrupedal locomotion \cite{rudin2022learning}, and quadrotor control \cite{hwangbo2017control}. There is limited research work in CT with multiple ground robots using deep RL. The research of \cite{zhang2020decentralized} presents a multi-robot CT approach using deep Q networks (DQN). Similarly, Manko et al. \cite{manko2018adaptive} presented multi-robot CT with DQN, relying on a path planning algorithm. Eoh et al. \cite{eoh2021cooperative} presents a curriculum-based deep RL approach for multi-robot CT.
Further, limited work exists on multi-drone CT with deep RL such as those of \cite{chen2023robust,li2021trajectory}. However, in these works, all the agents deploy RL, use cabled or flexible connections, are suitable for indoor environments with reliance on precise motion capture systems, and, most importantly, the uncertainties, such as time-varying CG, are not considered. This paper, on the other hand, addresses these gaps.\\
This paper uses a leader-follower approach to address the problem of time-varying CG in multi-drone CT systems. It should be noted that, we take motivation from \cite{barawkar2019cooperative}, to use a PID based leader as it can inherently handle CG uncertainties. The follower drone implements a Soft Actor-Critic (SAC) based RL algorithm using local measurements (follower's position and velocity) and minimal leader information (leader's error from goal position) to control the effect of time-varying CG. It should be noted that an RL follower drone is required to ascertain the position, velocity, and goal location of the object/leader to provide stable system performance in the presence of system uncertainties, such as moving CG. This is evident from prior literature \cite{kawasaki2003adaptive,culbertson2018decentralized,barawkar2024decentralized}. It should be noted that inter-drone (except with the leader or the object) communication can still be avoided as only a single GPS sensor can be used to relay the object's/leader's information to all the drones. TNPG, TRPO, DDPG, and SAC form a few of the relevant RL algorithms used in literature for deep RL \cite{duan2016benchmarking,haarnoja2018soft}. 
This paper uses Soft Actor-Critic (SAC) \cite{haarnoja2018soft} to implement deep RL on the follower drone. SAC performed better over other algorithms \cite{haklidir2021guided,haarnoja2018composable} and had better exploration capabilities. Consequently, it was chosen as a starting point for this research, and other methods will be explored in future work. 

This paper's main contribution is developing the decentralized controller for the follower drone based on deep RL using SAC, which can handle the object's time-varying CG. This is the first work to implement deep RL for the proposed problem. Additionally, gravity, attitude coupling due to rigid connections, and coupled nonlinear drone dynamics add to the challenges in stabilization. We also compare the proposed RL-based control scheme with an adaptive controller \cite{barawkar2024decentralized} previously designed for the problem of time-varying CG. The presented results validate the effectiveness of the proposed controller over an adaptive controller in terms of better waypoint navigation. The proposed RL-based SAC controller has also been demonstrated to perform favorably for faster moving CG speeds and changes in the payload or mass of the object. Simulation results validate the effectiveness of the proposed method. Finally, a preliminary experimental demonstration of a two-drone CT system with ball balance mounted on a test stand is also conducted to illustrate the developed results. In summary, a novel RL-based decentralized controller that does not utilize force sensors and primarily relies on local measurements and minimal leader information for the follower drones has been developed and validated using simulations and experiments.

\begin{figure}[]
\centering
\includegraphics[width = 8.5 cm, height = 8cm]{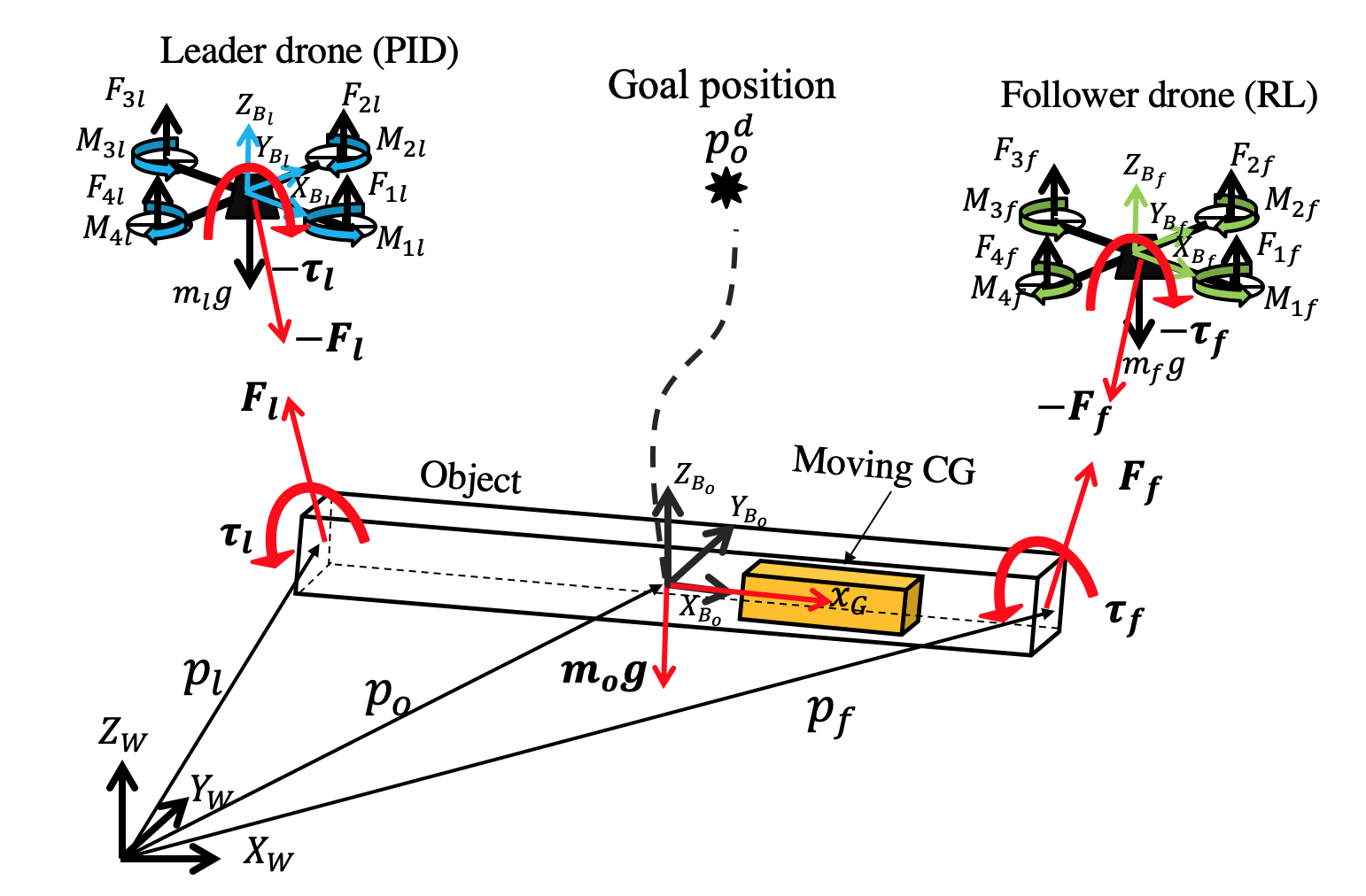}
\caption{Free body diagram of the entire system used for package delivery (showing moving CG).}
\label{fbdd}
\end{figure}

\section{SYSTEM DYNAMICS}
The dynamics of the leader-follower two-drone CT system have been discussed in \cite{barawkar2017admittance}. For the sake of clarity, they are briefly presented here. Refer to Fig. \ref{fbdd} showing the two-drone CT system for package delivery. $X_W$, $Y_W$ and $Z_W$ are the axes in the world frame $W$. $X_{B}$, $Y_{B}$ and $Z_{B}$ are the body frame axes with suffices $l$, $f$, and $o$ for indicating the leader drone, the follower drone, and the object being transported. It should be noted that the body frames of the leader drone, the follower drone, and the object are aligned since they share the same orientation due to the use of rigid connections in the system. We now write the equations of motion of the leader drone as follows,
\begin{equation}
F_l=R\begin{bmatrix}
0 & 0 &  F^t_l
\end{bmatrix}^T-\begin{bmatrix}
0 &
0 & 
m_lg
\end{bmatrix}^T-m_l\ddot{p}_l 
\end{equation}
\begin{equation}
\tau_l=\begin{bmatrix}
L_l*(F_{2l}-F_{4l})\\ 
L_l*(F_{3l}-F_{1l})\\ 
M_{1l}-M_{2l}+M_{3l}-M_{4l}
\end{bmatrix}-\omega\times I_l\omega-I_l \dot{\omega}
\end{equation}
where $F_l$ is the contact interaction force acting on the object at the point of contact between the leader drone and the object. $F^t_l$ is the sum of all the rotor forces of the leader drone. $m_l$ and $p_l$ and $R$ are the mass of the leader drone, the position of the leader drone in frame $W$, and the $Z-X-Y$ rotation matrix from the body frame $B_o$ (or $B_l$/$B_f$) to the world frame $W$, respectively. Refer to \cite{barawkar2017admittance} for the rotation matrix. Here, $\tau_l$ is the contact torque at the point of contact of the leader drone with the object, $L_l$ denotes the distance between the drone's geometric center and the leader's rotor center, $F_{il}$ and $M_{il}$ are the rotor forces and moments generated by the $i^{th}$ rotor of the leader drone, $I_l$ is the mass moment of inertia of the leader drone, and $\omega$ denotes the angular velocity of the system. Here, $\omega=[\dot{\phi},\dot{\theta}, \dot{\psi}]^T$, due to the small angle approximation \cite{barawkar2017admittance}, where $\phi$, $\theta$ and $\psi$ are the roll, pitch and yaw angles denoting the orientation of the system. Similar equations of motion can be written for the follower drone where $F_f$, $F^t_f$, $m_f$, $p_f$, $\tau_f$, denote the contact force acting on the object due to the follower drone, the total thrust, mass, position and contact torque of the follower drone, $F_{if}$ and $M_{if}$ are the forces and moments produced by the $i^{th}$ rotor of the follower drone, and  $I_f$ is the moment of inertia of the follower drone. We now write the equations of motion of the object (linear and rotational) as follows,
\begin{equation}
\ m_o\ddot{p}_o=F_l+F_f-\begin{bmatrix}
0 &
0 & 
m_o g
\end{bmatrix}^T
\end{equation}
\begin{equation}
\ I_o \dot{\omega}=\tau_l+\tau_f-(\omega\times I_o\omega) +(\mathbf{r_l}\times F_l)+(\mathbf{r_f}\times F_f)
\label{eom-rp}
\end{equation}
\noindent where, $\mathbf{r_l}= p_l-p_o$ and $\mathbf{r_f}=p_f-p_o$. $m_o$ and $I_o$ are the object's mass and moment of inertia, and $p_o$ is the position of the object's geometric center. The rigid body equation is then written as $\ddot{p}_l=\ddot{p}_o+(\dot{\omega}\times \mathbf{r_l})+(\omega\times (\omega\times \mathbf{r_l}))$. A similar equation can be written to compute the linear acceleration of the follower drone. This detailed dynamic model was used to develop a realistic simulation framework for the custom environment used in this paper to train the deep RL follower - PID leader drone CT.

\section{APPROACH}
To implement a multi-drone CT task, a leader-follower approach is used. This section describes the control strategies used for the leader and follower drones, respectively. 

\subsection{Leader control}
\noindent Similar to \cite{barawkar2019cooperative}, the leader drone implements a PID controller to navigate the object (or the leader) from the origin to the desired goal position $p_o^d$ (see Fig. \ref{fbdd}) of the object as,
\begin{equation}
\ddot{p}_l^d = k_p e_o + k_d \dot{e}_o + k_{I}\int e_o dt
\label{pid}
\end{equation}
where, $k_p$, $k_d$ and $k_I$ are the proportional, derivative and integral gains and $e_o= p_o^d - p_o$. $\ddot{p}_l^d$ constitutes the desired linear acceleration of the leader drone for goal point navigation. From the above equation, the desired roll and pitch angles for the leader drone are computed by linearization of the dynamic equations as follows,
\begin{align*}
\phi_l^d =\frac{(\ddot{p}_{xl}^d\sin\psi^d_l-\ddot{p}_{yl}^d\cos\psi^d_l)}{g} &\; \theta_l^d=\frac{(\ddot{p}_{xl}^d\cos\psi^d_l+ \ddot{p}_{yl}^d\sin\psi^d_l)}{g}
\label{e19}
\end{align*}
where $\ddot{p}_{xl}^d$, $\ddot{p}_{yl}^d$ and $\ddot{p}_{zl}^d$ denote the components of the desired acceleration of the leader drone along $X_W$, $Y_W$ and $Z_W$ axes. The control action $u_l = (\phi_l^d,\theta_l^d, \ddot{p}_{zl}^d)$ is then commanded to the attitude controller of the leader drone, which computes the desired rotor speeds of the leader drone based on the above-generated control action $u_l$. Refer to \cite{barawkar2017admittance} for details.

\subsection{Follower control using deep RL}
This sub-section describes the deep RL architecture used to control the follower drone. We first describe the notations used in the context of RL, followed by a brief explanation of SAC (for completeness of the paper) and the reward formulation and control of the follower as follows:\\

\noindent\textit{1) Notations used in the context of RL:}\\
Consider policy learning in continuous action spaces with a Markov Decision Process (MDP) of tuple $(S, \mathcal{A}, \mathfrak{p}, r)$. Where $S$ and $\mathcal{A}$ are the continuous state and action spaces, $\mathfrak{p}$ is the probability distribution of next state $s_{t+1}$ given the current state $s_t$ and the action $a_t$ and $r_t$ is the bounded reward of each step or transition. A policy is defined by $\pi(a_t|s_t)$. Further, $\rho_\pi (s_t)$ and $\rho_\pi (s_t, a_t)$ represent the state and state-action marginals of the trajectory distribution created by the policy $\pi$ \cite{haarnoja2018soft}.\\

\noindent\textit{2) Soft Actor-Critic (SAC) - Brief Overview:}\\
SAC is an off-policy deep RL algorithm to address continuous and discrete control problems with better exploration capabilities \cite{tang2021novel}. SAC provides a trade-off between exploitation and exploration with entropy regularization. The increase in entropy results in higher exploration. Moreover, SAC can avoid bad local optimum as well \cite{haklidir2021guided}. The maximum entropy RL favors stochastic policies by amplifying the objective with the expected entropy of policy over $\rho_\pi (s_t)$. Thus, the objective function of SAC consisting of reward and an additional entropy term ($\mathcal{H}$) is $J(\pi) = \sum_{t=0}^{T}\mathbb{E}_{(s_t,a_t)\sim \rho_\pi}\left [ r(s_t,a_t) + \alpha\mathcal{H}(\pi(\cdot |s_t)) \right ]$. $\alpha$ denotes the hyperparameter of the temperature coefficient controlling the optimal policy stochasticity or focus of entropy. Now, we define the three networks that are utilized in the SAC algorithm. They are the state value function $V$, a soft $Q$ function, and a policy function $\pi$ with parameter suffices as $\Psi$, $\Theta$, and $\Phi$. The value and the $Q$ networks are trained by minimizing the squared residual error and the soft Bellman residual as,
\begin{equation}
\begin{split}
J_V(\Psi)= \mathbb{E}_{(s_t)\sim \mathcal{D}}\Big[\frac{1}{2}\Big(V_\Psi(s_t) - \mathbb{E}_{a_t\sim\pi_\Phi}\Big[ Q_\Theta (s_t,a_t)\\ - log \pi_\Phi(a_t|s_t) \Big]\Big)^2 \Big]
\end{split}
\end{equation}
\begin{equation}
\begin{split}
J_Q(\Theta)= \mathbb{E}_{(s_t,a_t)\sim \mathcal{D}}\Big[ \frac{1}{2}\Big(Q_\Theta(s_t,a_t)-\Big(r(s_t,a_t)\\+\gamma\mathbb{E}_{s_{t+1}}\Big [ V_{\bar{\Psi}}(s_{t+1}) \Big]\Big) \Big)^2 \Big]
\end{split} 
\end{equation}
$\mathcal{D}$ is the distribution of the replay buffer, and the actions are sampled with respect to the current policy. Further, minimization of the expected Kullback-Leibler (KL) divergence is performed to estimate the policy parameters as follows, 
\begin{equation}
   J_\pi(\Phi) = \mathbb{E}_{s_t\sim \mathcal{D}}\Big[D_{KL}\Big(\pi_\Phi(\cdot | s_t)||\frac{exp(Q_\Theta (s_t,\cdot))}{Z_\Theta(s_t)}\Big)\Big] 
\end{equation}
A neural transformation is then utilized to reparameterize the policy; see \cite{haklidir2021guided} for more details. \\
\begin{figure}[]
\centering
\includegraphics[width=6 cm, height=4cm]{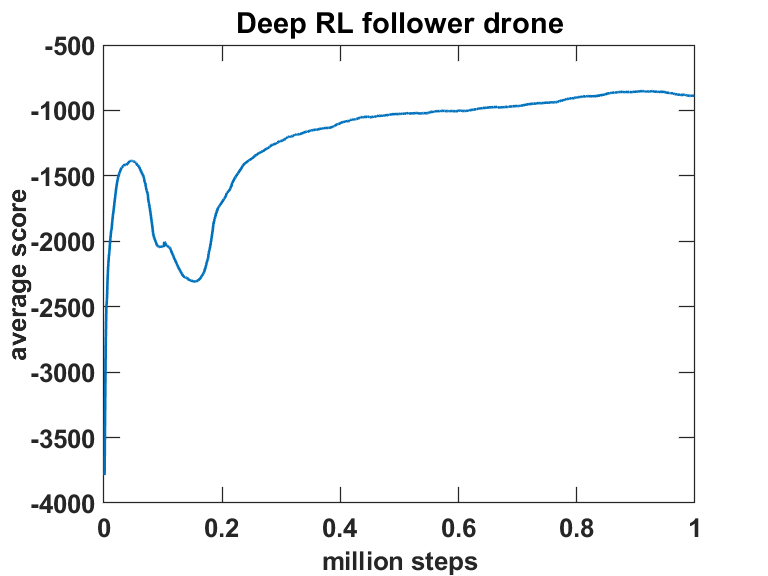}
\caption{Performance of SAC implemented on the follower.}
\label{rew}
\end{figure}

\noindent\textit{3) Reward formulation and control of the follower drone:}\\
The multi-drone CT task is started at $t=0$ time step and terminated after $1000$ steps or if the distance of the object/payload from the goal position becomes greater than a certain threshold value $d^{term}$. Each such event is called an episode. We run multiple episodes ($1000$ in our simulation setup) to train the SAC agent. We now define the states and actions and formulate the reward function for the RL follower drone.\\
Let $s_{f,t}$ denote the state of the RL follower drone. The state or observation or input of the RL follower drone is $s_{f,t}=[p_f, \eta, \dot{p}_f, \dot{\eta}, e_o, \dot{e_o}]$ consisting of local measurements of follower drone's position, velocity (both linear and angular) and minimal object information. The continuous output or the action produced by the RL-based follower drone consists of its desired control action viz. $a_{f,t}=u_f=(\phi^d_f,\theta_f^d, \ddot{p}^d_{zf})$ (desired roll, pitch and vertical acceleration of the follower drone) at the time step $t$. After the RL follower takes this action, the system transitions to the next time step with the next state $s_{t+1}$ receiving a reward $r_f(s_{f,t},a_{f,t})$ for the particular action it took. The reward is then formulated as, 
\begin{equation*}
    r_f(s_{f,t},a_{f,t}) = -10\left\| \ddot{p}^d_{zl}- \ddot{p}^d_{zf}\right\|-10\left\| \phi^d_l - \phi_f^d\right\| - 10\left\|\theta^d_l - \theta_f^d \right\|
\end{equation*}
Note that this reward function ensures that the RL follower drone agent mimics the leader drone's control action without inter-drone communication. This ensures effective waypoint navigation and balance of moving CG. This entire work assumes that the leader is perfect in terms of stable waypoint navigation, which will, in turn, result in a stable overall performance of the follower drone, thus ensuring effective PID leader - RL follower cooperation. At each time step, the desired control action of the follower drone ($a_{f,t}=u_f=(\phi^d_f,\theta_f^d, \ddot{p}^d_{zf})$) is then commanded
to the follower's attitude controller. Additionally, without the loss of the ability of the follower to cooperate with the leader, the desired yaw angle for the follower drone is assumed to be $\psi^d_f = 0$.
\begin{figure}[t]
\centering
\includegraphics[width = 8 cm, height = 8cm]{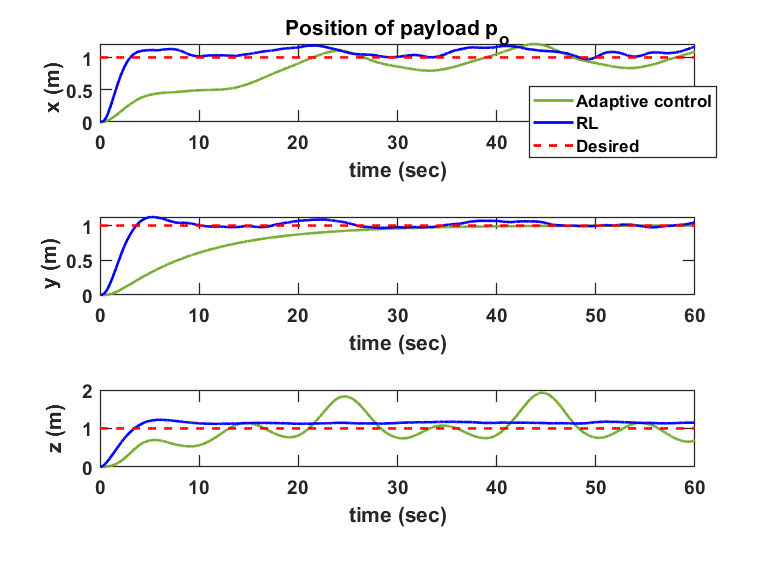}
\caption{Position of the geometric center of payload or object being transported by two drones for adaptive and RL controllers implemented on follower drone.}
\label{pos_rl_adap}
\end{figure}


\section{RESULTS}
\subsection{Simulations}
We implemented a unique custom RL environment to simulate the multi-drone CT system with moving CG using Python, our SAC Agent, and the OpenAI Gym framework. The RL environment utilized the combined payload and drone dynamics to simulate the system using rigid connections between the drones and the payload/object. The following parameters were used, learning rate for SAC RL agent (actor, value and critic networks) - $0.0003$, the discount factor of $\gamma = 0.99$, the entropy coefficient - $\alpha = 0.3$, replay size - $1000000$, two layer neural networks were used for actor, critic and value networks with $256$ neurons in each layer, batch size - $256$, $d^{term} = 2.5m$, $k_p = [0.5,0.5,0.5]^T$, $L = 0.12m$, length of the object $len=0.34m$, $m_o = 0.2kg$, $m_l=m_f = 1kg$, $k_d = [1,1,1]^T$ and $k_I = [0,0,0]^T$.

\begin{figure}[htbp]
\centering
\includegraphics[width=8 cm, height=8cm]{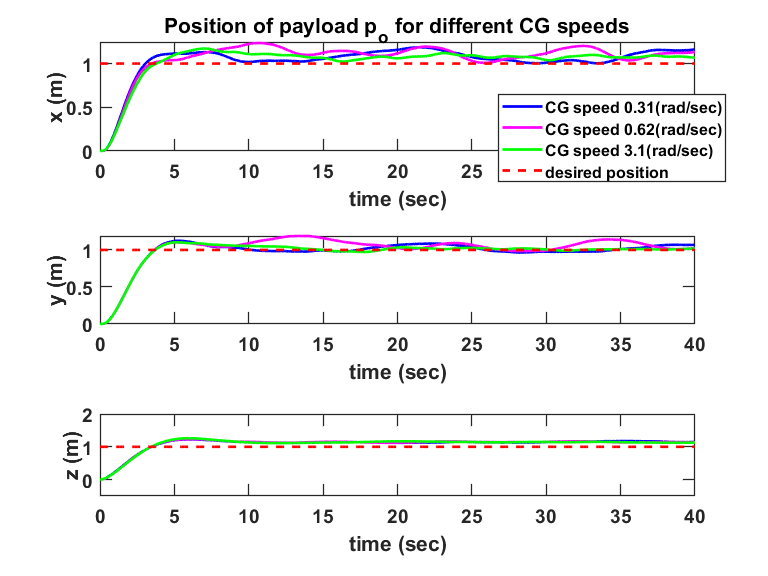}
\caption{Position of the geometric center of payload $p_o$ using deep RL follower and PID leader tested for different speeds of CG.}
\label{cg_speed}
\end{figure}

\begin{figure}[htbp]
\centering
\includegraphics[width=8 cm, height=8cm]{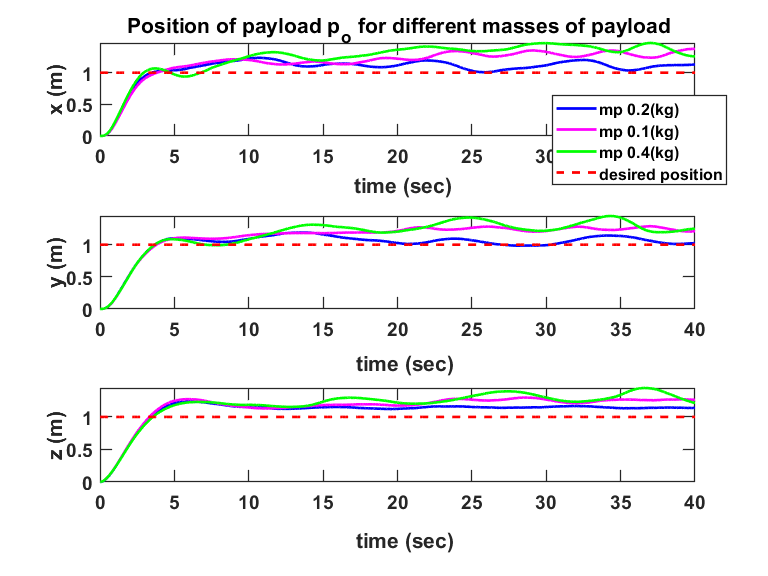}
\caption{Position of the geometric center of payload $p_o$ using deep RL follower and PID leader tested for the different mass of the object being transported.}
\label{mass_add_drop}
\end{figure}

The SAC RL follower was trained for over a million time steps, and the plot of average reward is shown in Fig. \ref{rew}, indicating the effectiveness of the trained controller. Figure \ref{pos_rl_adap} shows the position of the object's geometric center. The CG was made to move along the $X_{B_o}$ axis with an oscillatory angular velocity of $0.31 rad/sec$ and was trained for this CG speed and constant mass of the object of $m_o=0.2kg$. The leader control consists of a traditional PID controller to reach the goal position, while the follower uses the proposed deep RL-based controller. The overall system results of the proposed RL-based controller are compared with an adaptive controller \cite{barawkar2024decentralized} (designed to estimate both the CG and the rate of change of CG). Thus, the deep RL follower-PID leader system is compared with the adaptive follower-leader system in the above-indicated figures. It can be observed from the Figure \ref{pos_rl_adap} that the proposed deep RL-PID system works better than an adaptive system with lesser oscillations, peaks, and better settling time for the proposed method. This is expected since the adaptive controller attempts to estimate the CG and its rate and then feeds it to the controller, where certain delays in estimation (of CG and its rate) can result in oscillatory and spiky behavior, as seen in the results. See reference \cite{barawkar2024decentralized} for more details. \\
We tested the RL controller for different CG speeds and object mass variations. Note that the RL controller was trained for slow CG and constant mass as indicated earlier. As seen in Figures \ref{cg_speed} and \ref{mass_add_drop} that the proposed RL controller and the overall system can stably tolerate CG speed (slow to fast) and object mass variations (light and heavy). 

\subsection{Experiments}
\begin{figure*}[htbp]
\centering
\includegraphics[width=14 cm, height=4cm]{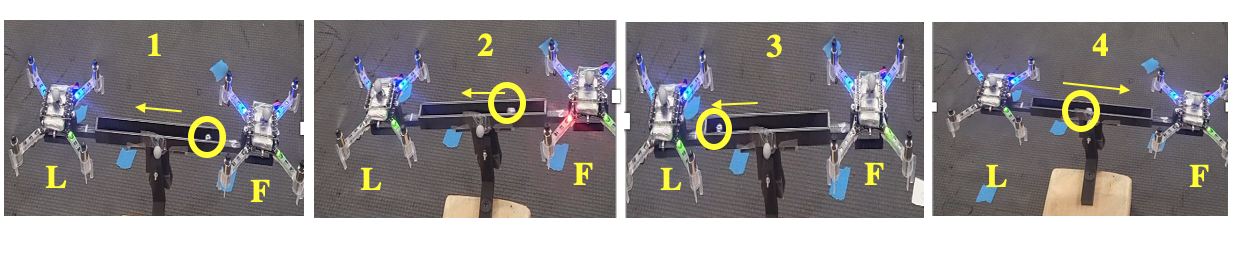}
\caption{Experimental setup showing two Crazyflies attached to a stick with a moving ball (symbolizing moving CG). The images are numbered from 1 to 4 for readability. L and F denote the PID leader and the RL follower, respectively. The yellow circle shows the position of the ball, and the yellow arrow shows the direction of motion of the ball. }
\label{exp_set}
\end{figure*}

Refer to Fig. \ref{exp_set} showing the experimental setup of two Crazyflie 2.1 drones attached rigidly to a stick with a moving ball (symbolizing moving CG). Roll, pitch, and thrust commands were sent to the Crazyflies for the leader and the follower drones. 
It should be noted that Crazyflie nano drones accept thrust commands from the range of $10000(0\%)$ to $60000(100\%)$. Thus, the thrust or the vertical $Z_W$ axis controller was designed as, 
\begin{align}
thrust_l = thr_{hov} + k_{pz}\ddot{p}_{zl}^d & \;\;\;thrust_f = thr_{hov} + k_{pz}\ddot{p}_{zf}^d
\end{align}
where $thr_{hov}$ and $k_{pz}$ are the hovering thrust and the vertical control proportional constant. 

\begin{figure}[htbp]
\centering
\includegraphics[width=7 cm, height=7cm]{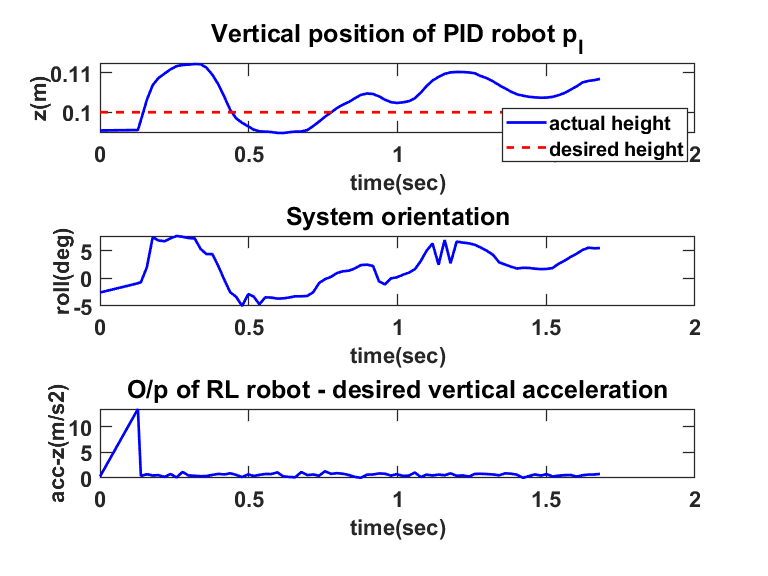}
\caption{Vertical position of the leader drone $p_{zl}$, roll angle of the system and the desired vertical acceleration of the follower drone using deep RL.}
\label{experi}
\end{figure}

The control actions of the leader and the follower drones are then commanded to the attitude controller of the Crazyflie platforms. Thus $u_l=(\phi_l^d,\theta_l^d,thrust_l)$ is commanded by the PID leader Crazyflie drone and $u_f=(\phi_f^d,\theta_f^d,thrust_f)$ is commanded by the RL follower Crazyflie. Note that $\ddot{p}_{zl}^d$ and $\phi_l^d$ are computed by the leader's PID controller while $\ddot{p}_{zf}^d$ and $\phi_f^d$ are derived from the SAC based RL controller of the follower drone (trained in simulations). We do not command desired pitch angles to the drones since the test stand is restricted along the pitch axis and only allowed to roll and change vertical heights. Hence, $\theta_l^d = \theta_f^d = 0$. The parameters used for experiments are, $k_p = [1,1,0.1]^T$, $k_d= [2,2,5]^T$, $k_I= [0.2,0.1,0.1]^T$, $k_{pz}=1000$, $thr_{hov}=46000$. It should be noted that since $p_o$ is restricted to moving as can be seen in Fig. \ref{exp_set} of the test stand, we use $e_l=p_l^d-p_l$ for the leader's PID controller instead of $e_o$ in equation \ref{pid}, where, $p_l^d=[0,0,0.08]^T$ is the desired position of the leader drone using PID controller. It should be noted that this does not affect the system performance, given the leader's task to navigate towards a goal position. It can be observed from Figures \ref{experi} and \ref{exp_set} that the two-drone CT system effectively balances the moving CG (or ball) and tries to maintain zero roll angle (to balance the ball). 


\section{CONCLUSION}
This paper presented a leader-follower approach for controlling a multi-drone CT system in the presence of time-varying CG. The leader utilized a PID controller, while the follower was controlled using a deep RL SAC controller. The follower used its local position and velocity measurements and minimal leader/object information to cooperatively transport an object with the leader in the presence of moving CG. The proposed method was shown to be effective when compared with an adaptive controller based balancing framework, wherein the proposed strategy also tolerated higher CG speeds and different object mass variations. Preliminary experimental results of ball balance on a stick using two drones also demonstrated the efficacy of the proposed method.
Future work consists of experimental validation of actual flight (not test stand) of such systems with moving CG.








\bibliographystyle{IEEEtran}
\bibliography{IEEEabrv,shr}

\end{document}